\documentclass[10pt,twocolumn,letterpaper]{article}

\usepackage{wacv}
\usepackage{times}
\usepackage{epsfig}
\usepackage{graphicx}
\usepackage{amsmath}
\usepackage{amssymb}

\usepackage{subfigure}

\usepackage{color}
\definecolor{blue}{rgb}{0,0,1}
\definecolor{red}{rgb}{1,0,0}
\definecolor{green}{rgb}{0,.6,0}
\definecolor{black}{rgb}{0,0,0}
\wacvfinalcopy % *** Uncomment this line for the final submission

\def\wacvPaperID{385} % *** Enter the wacv Paper ID here

\ifwacvfinal\fi
\begin{document}

%%%%%%%%% TITLE
\title{Semi-dense Stereo Matching using Dual CNNs}

% Authors at the same institution
%\author{First Author \hspace{2cm} Second Author \\
% Authors at different institutions
\author{Wendong Mao$^1$, Mingjie Wang$^1$, Jun Zhou$^2$, Minglun Gong$^{1*}$\\
$^1$Memorial University of Newfoundland\\
$^2$Dalian University of Technology\\
%{\tt\small wm0330, mingjiew, gong@mun.ca}
%\and
}

%  wm0330@mun.ca
%Mingjie Wang mingjiew@mun.ca
%Jun Zhou  junz@mun.ca
%Minglun Gong* gong@mun.ca

\maketitle
\ifwacvfinal\thispagestyle{empty}\fi

%%%%%%%%% ABSTRACT
\begin{abstract}
A robust solution for semi-dense stereo matching is presented. It utilizes two CNN models for computing stereo matching cost and performing confidence-based filtering, respectively. Compared to existing CNNs-based matching cost generation approaches, our method feeds additional global information into the network so that the learned model can better handle challenging cases, such as lighting changes and lack of textures.
Through utilizing non-parametric transforms, our method is also more self-reliant than most existing semi-dense stereo approaches, which rely highly on the adjustment of parameters. The experimental results based on Middlebury Stereo dataset demonstrate that the proposed approach outperforms the state-of-the-art semi-dense stereo approaches.

\end{abstract}

\section{Introduction}

Humans rely on binocular vision to perceive 3D environments. Even though it is a passive system, our brains can still estimate 3D information more rapidly and robustly than many active or passive sensors that have been developed.  One of the reasons is that brains can utilize prior knowledge to understand the scene and to infer the most reasonable depth hypothesis even when the visual cues are lacking. Recent advances in machine learning have shown that the brain's discrimination power can be mimicked using deep convolutional neural networks (CNNs). Hence, one has to wonder how CNNs can be used to enhance traditional stereo matching algorithms.

Approaches have been proposed for generating matching cost volumes (a.k.a. disparity space images) using CNNs \cite{park2017look,ye2017efficient,zbontar2016stereo}. While inspiring results are generated, these existing approaches are not robust enough for handling challenging and ambiguous cases, such as lighting changes and lack of textures.  Heuristically defined post-processing steps are often applied to correct mismatches.  Our hypothesis is that the performance of CNNs can be noticeably improved if more information is fed into the network. Hence, instead of trying to correct mismatches as post-processing, we introduce, in the pre-processing step, image transforms that are robust against lighting changes and can add distinguishable patterns to textureless areas. The output of these transforms are used as additional information channels,
together with grayscale images, for training a matching CNN model.

%\ml{which, together with the color channels, are used for training a matching CNN model.} 
The experimental results show that the model learned can effectively separate correct stereo matches from mismatches so that accurate disparity maps can be generated using the simplest Winner-Take-All (WTA) optimization.

%On the other hand, l
Learning-based approaches were also proposed to compute confidence measure for generated disparity values so that mismatches can be filtered out \cite{cheng2018learning, seki2016, ye2017efficient}. Following this idea, a second CNN model is designed to evaluate the disparity map generated through WTA.  Trained with only one input image and the disparity map, this evaluation CNN model can effectively filter out mismatches and produce accurate semi-dense (a.k.a. sparse) disparity maps.

\begin{figure*}[t]
\begin{center}
   \includegraphics[width=0.80\linewidth]{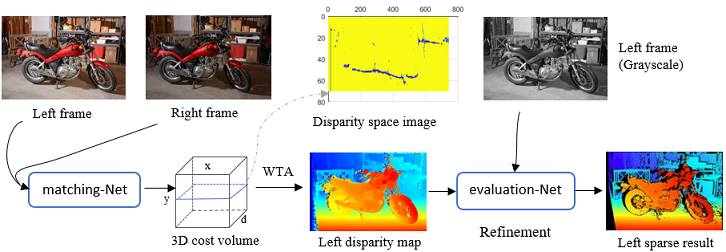} 
\end{center}
\caption{Semi-dense stereo matching pipeline. Given a pair of rectified images, how well a pair of image patches match is evaluated using \emph{a matching-CNN model}. The results form a matching cost volume, from which a disparity map is generated using simple WTA optimization. Finally, an \emph{evaluation-CNN model} is applied to filter out mismatches. } %\mlc{Can u add a box representing transformations?}
\label{fig:pipeline}
\end{figure*}

Figure \ref{fig:pipeline} shows the pipeline of the whole process. Since both matching cost generation and disparity confidence evaluation are performed using learning-based approach, the algorithm contains very few handcrafted parameters. The experiment results on Middleburry 2014 stereo dataset \cite{scharstein2014high} demonstrate that the present dual-CNN algorithm outperforms most existing sparse stereo techniques. 
%. Depth sensation is the corresponding term for animals    it is surprising how fast $f$ $l$ $p$

\section{Related Work}

Stereo matching algorithms can be briefly categorized into two classes: dense and sparse stereo matching \cite{lazaros2008review, scharstein2002taxonomy}. Dense stereo matching algorithms assign disparity values to all pixels, whereas sparse matching approaches only output disparities for pixels with sufficient visual cues. A typical pipeline implemented in most dense and sparse stereo matching algorithms consists of matching cost computation, cost aggregation, disparity computation and optimization, and refinement steps \cite{scharstein2002taxonomy}. Here, we focus our discussion on sparse stereo matching. 
%the latter class.

An early work of sparse stereo matching was implemented by calculating disparities for distinctive points first and gradually generating disparity values for the remaining pixels \cite{manduchi1999distinctiveness}. Graph cuts was later introduced in \cite{veksler2003extracting} to detect textured areas as an alternative to unambiguous points and generate corresponding semi-dense results. By design, their approach can filter out mismatches caused by lack of textures, but not by occlusions, etc. This limitation is addressed in Semi-Global Matching (SGM) \cite{hirschmuller2008stereo}, in which multiple 1D constraints were used to generate accurate semi-dense results based on peak removal and consistency checks. Gong and Yang \cite{gong2003consistency} proposed a reliability measure to detect potential mismatches from disparity maps generated using Dynamic Programing (DP). This work was later extended and implemented on graphics hardware for real-time performance \cite{gong2005near}. 
%Various techniques were proposed more recently to further enhance the performances of sparse stereo matching. A 2D mesh via triangulation supported by a set of sparse matching points was proposed by Geiger et al. \cite{geiger2010efficient} to reduce matching ambiguities. Following the idea of triangulation, a set of line segments and support points was introduced in \cite{jellal2017ls} to generate a more informative triangulation mesh which can better handle depth discontinuities. 
%Furthermore, 
Psota et al. \cite{psota2015map} utilized Hidden Markov Trees (HMT) to create minimum spanning trees based on color information which allows aggregated costs to be passed along the tree branches, and the isolated mismatches were later moved by median filtering.

Generally, the above algorithms utilize additional constraints in the cost aggregation or/and disparity computation step to improve the accuracy of sparse stereo matching. Instead of designing new constraints or assumptions, we train CNNs for both generating aggregated cost volumes and detecting potential mismatches. The disparity computation step, on the other hand, is performed using the simplest WTA approach.

\subsection{Stereo Matching Cost}

Traditionally, sum of absolute differences (SAD), sum of squared differences (SSD), and normalized cross-correlation (NCC) had been commonly used for calculating and aggregating matching costs \cite{scharstein2002taxonomy}. These window-based matching techniques that rely on the local intensity values may not behave well near discontinuities in disparity. Zabih and Woodfil \cite{Zabih} therefore proposed two non-parametric local transforms, referred to as rank and census transforms, to address the correspondences at the boundaries of objects. A recent attempt tried to combine different window-based matching techniques for stereo matching~\cite{batsos2018cbmv}.

In the past few years, various works were proposed to generate matching cost volumes that can better differentiate correct matches from mismatches. The most encouraging direction is using ground-truth data %\cite{simonyan2014learning}
\cite{han2015matchnet, simonyan2014learning, trzcinski2015learning, zagoruyko2015learning}
to train various neural networks to learn local image features. More recent works mostly opted for CNNs trained by ground-truth data 
%\mlc{does this applies to all cited papers above?} 
to predict the likeness for each potential match based on fixed windows as in \cite{zbontar2016stereo}. The matching costs were set using the output of CNNs directly.

Due to the successful practice of using CNNs, stereo matching algorithms have been progressively improved over the past three years.  Zhang et al. \cite{zhang2017robust} used CNNs and SGM to generate initial disparity maps and further combine Left Right Difference (LRD) \cite{hu2012quantitative} with disparity distance regarding local planes to perform confidence check. In addition, they adopted segmentation and surface normal within the post-processing to enhance the reliability of disparity estimation. To fully utilize the ability of CNNs in terms of feature extraction, Park and Lee \cite{park2017look} proposed a revised CNN model based on a large pooling window between convolutional layers for wider receptive fields to compute the matching cost, and they performed similar post-processing pipeline introduced in \cite{zbontar2016stereo}. Another model revision, similar to Park and Lee's work \cite{park2017look},  was introduced by Ye et al. \cite{ye2017efficient}, which used a multi-size and multi-layer pooling scheme to take wider neighboring information into consideration. Moreover, a disparity refinement CNN model was later demonstrated in their post-processing to blend the optimal and suboptimal disparity values. Both the above revisions presented solid results in image areas with low or devoid of texture, disparity discontinuities and occlusions. 

Attempts were also made to train end-to-end deep learning architectures for predicting disparity maps from input images directly, without the needs of explicitly computing the matching cost volume~\cite{chang2018pyramid, kendall2017end, pang2017cascade}. As a result, these end-to-end models are efficient but require larger amount of GPU memory than the previous patch-based approaches. More importantly, these models were often trained on stereo datasets with specific image resolutions and disparity ranges and hence, cannot be applied to other input data. They also restrict the feasibility of training CNNs to concurrently preserve geometric and semantic similarity proposed in \cite{choy2016universal, schmidt2017self, weerasekera2017learning}.

%The aforementioned approaches all trained CCNs on local image patches regardless size difference, but there were still a few attempts training CCNs in a global way. 

\subsection{Confidence Measure}

Once dense disparity results are generated, confidence measures can be applied to filter out inaccurate disparity values in the disparity refinement step. Quantitative evaluations on traditional confidence measures were presented by Hu and Mordohai \cite{hu2012quantitative}, and the most recent review was given by Poggi et al. \cite{poggi2017quantitative}. We hereby mainly presented a few significant works. 

Haeusler et al. \cite{haeusler2013ensemble} proposed a random decision forest framework, which combines multivariate confidence measures to improve error detection. Park and Yoon \cite{park2015leveraging} suggested a regression forest framework for selecting effective confidence measures. Based on SGM \cite{hirschmuller2008stereo}, Poggi and Mattoccia \cite{poggi2016learning} used O(1) features and machine learning to implement an improved scanline aggregation strategy, which performs streaking detection on each path in \cite{hirschmuller2008stereo} to perform confidence measure. They further proposed using a CNN model to enforce local consistency on confidence maps \cite{poggi2017learning}. Recently CNNs have been applied to confidence measure. Our approach is similar to these recent works \cite{cheng2018learning, seki2016, ye2017efficient}, which compute confidences through training 2D CNNs on 2D image or/and disparity patches. A key difference is, however, that only the left image and its raw disparity map generated by WTA are used to train our confidence CNN model, whereas existing approaches require to generate both left and right disparity maps.

\section{Methodology}

We aim at developing a robust and learning-based stereo matching approach. We observed that, for many applications, it is more important to ensure the accuracy of output than to generate disparity values for all pixels.  Hence, we here focus on assigning disparity values only for pixels with sufficient visual cues.

As shown in Figure~\ref{fig:pipeline}, two CNN models, referred as matching-Net and evaluation-Net, are utilized in our approach: matching-Net is constructed as the substitution of matching cost computation and aggregation steps, and outputs matching similarity for each pixel pairs; evaluation-Net performs confidence measure on the raw disparity maps generated by WTA based on the similarity scores.

\begin{figure*}[t]
\begin{center}
   \includegraphics[width=0.995\linewidth]{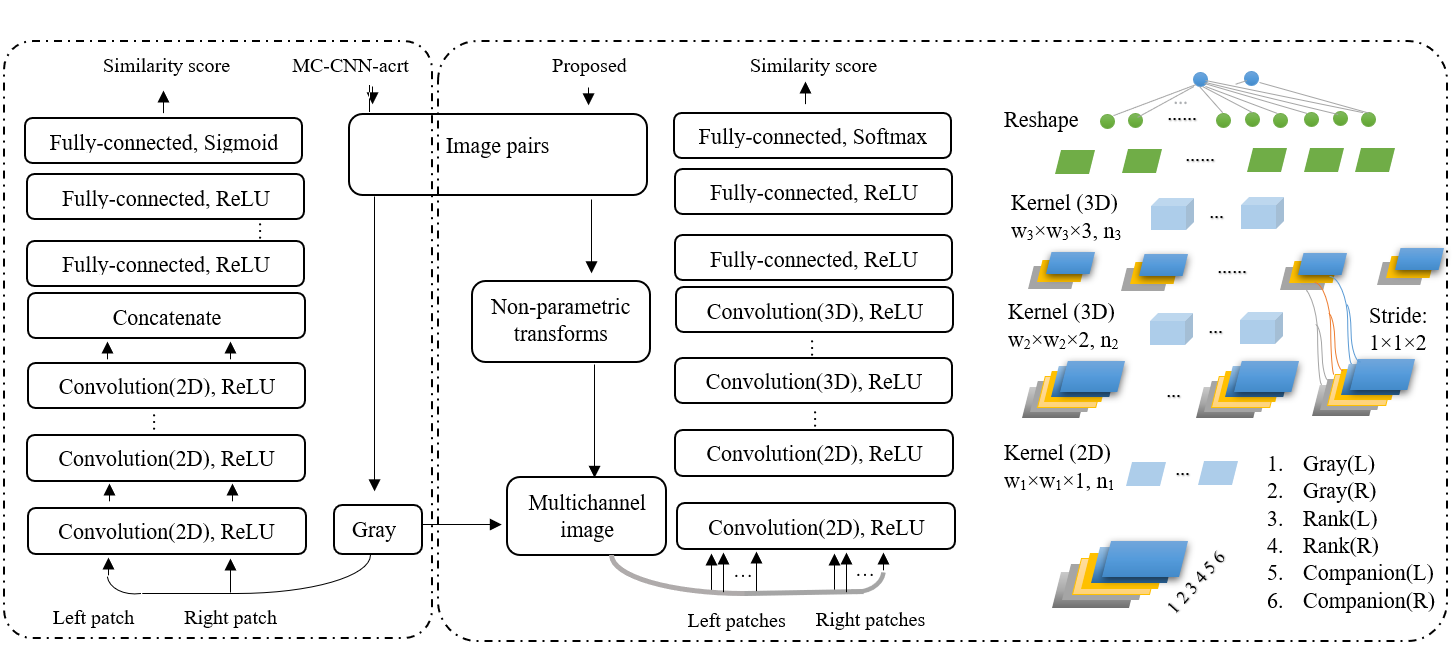} 
\end{center}
\caption{Comparison between the baseline architecture ``MC-CNN-arct" in \cite{zbontar2016stereo} and the proposed model matching-Net. The left and right image patches for the latter are selected from an image collection, which includes not only grayscale images, but also channels generated by non-parametric transforms. In addition, the concatenation operation is replaced by 3D convolution, which can separately group different transforms by adjusting stride size in the third dimension; see Section \ref{sec} for model configuration.}
\label{fig:net}
\end{figure*}

\subsection{Matching-Net}

Our matching-CNN model serves the same purpose as the ``MC-CNN-arct'' in \cite{zbontar2016stereo}, but there are several key differences; see Figure~\ref{fig:net}. First of all, we choose to feed the neural network with additional global information (i.e., results of non-parametric transformations) that are difficult to generate through convolutions. Secondly, 3D convolution networks are employed, which we found can improve the performance. It is worth noting that our approach is also different from other attempts to improve ``MC-CNN-arct'', which use very large image patches and multiple pooling sizes~\cite{park2017look,ye2017efficient}. These approaches require extensive amount of GPU memory, which limits their usage. In order to feed global information into the network trained on small patches, our strategy is to perform non-parametric transforms. 

%\subsubsection{Rank Transform.}
\subsubsection{Lighting Difference}

For robust stereo matching, lighting difference as an external factor cannot be neglected. To address this factor, ``MC-CNN-arct'' manually adjusted the brightness and contrast of image pairs to generate extra data for training. However, datasets with lighting difference may vary from one to another, making it hard to train a model that is robust against to all cases. 

Aiming for an approach with less human intervention, here we propose using rank transform to tolerate lighting variations between image pairs. As a non-parametric local transform, rank transform was first introduced by Zabih ~\cite{Zabih} to achieve better visual correspondence near discontinuities in disparity. This endows stereo algorithms based on rank transform with the capability to perform similarity estimation for image pairs with different lighting conditions. 

\begin{figure*}[t]
\centering
	\subfigure[Grayscale]{\includegraphics[width=0.19\linewidth]{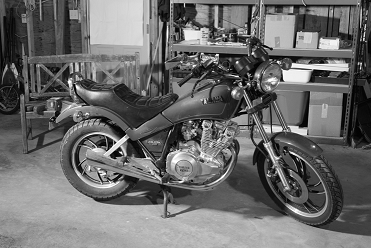}}
	\subfigure[$15 \times 15$]{\includegraphics[width=0.19\linewidth]{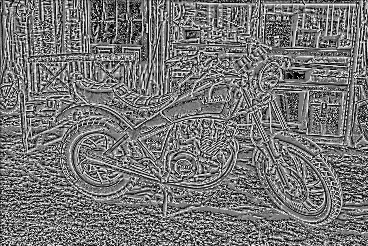}}
	\subfigure[$31 \times 31$]{\includegraphics[width=0.19\linewidth]{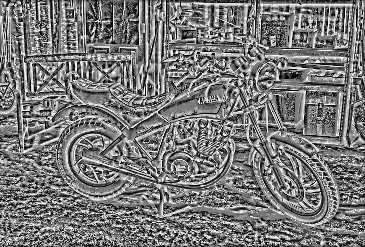}}
	\subfigure[$61 \times 61$]{\includegraphics[width=0.19\linewidth]{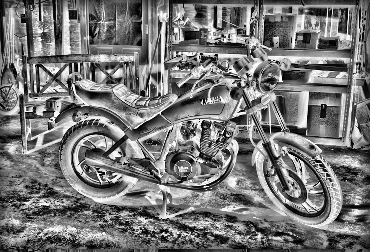}} \\
	\subfigure[Grayscale]{\includegraphics[width=0.19\linewidth]{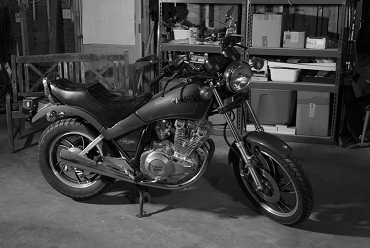}}
		\subfigure[$15 \times 15$]{\includegraphics[width=0.19\linewidth]{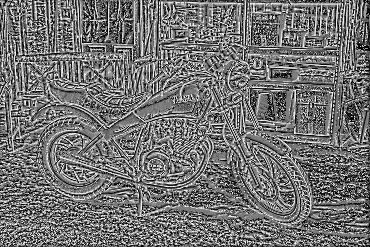}}
	\subfigure[$31 \times 31$]{\includegraphics[width=0.19\linewidth]{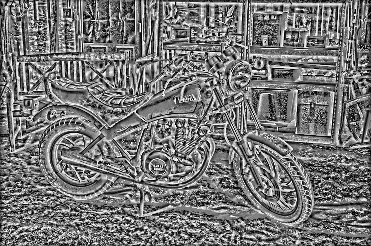}}
	\subfigure[$61 \times 61$]{\includegraphics[width=0.19\linewidth]{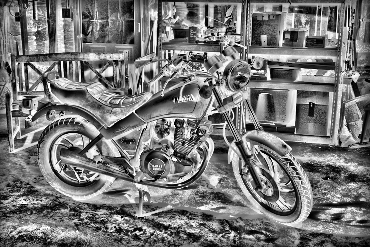}} 
\caption{Results of ``MotorE" dataset (left image on top row and right image on bottom) using rank transform under different neighborhood sizes. %The light difference between the image pair are diminished here. 
Lager windows generally lead to smoother results, but at the expense of losing subtle information.}
\label{fig:rank}
\end{figure*}

The rank transform $R(p)$ for pixel $p$ in image $I$ is computed as: 
\begin{equation}
R(p) = \frac{1}{|N_p|}\sum_{q \in N_p}{(I(q) > I(p) ? 1 : 0)} ~,
\label{eq:3_1}
\end{equation}
where $N_p$ is a set containing pixels within a square window centered at $p$. $|S|$ is the size of set S. Figure \ref{fig:rank} shows the results of rank transform under different window sizes.

\begin{figure*}[t]
\centering
	\subfigure[Grayscale]{\includegraphics[width=0.19\linewidth]{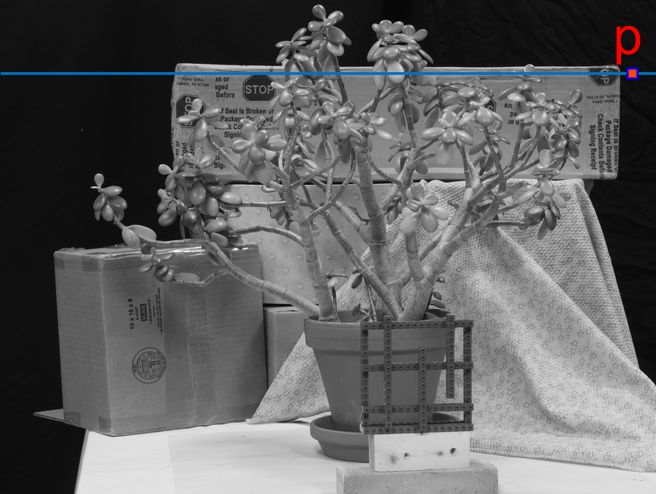}}
	\subfigure[$15 \times 15$]{\includegraphics[width=0.19\linewidth]{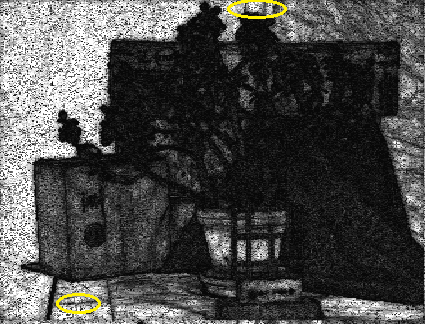}}
		\subfigure[$31 \times 31$]{\includegraphics[width=0.19\linewidth]{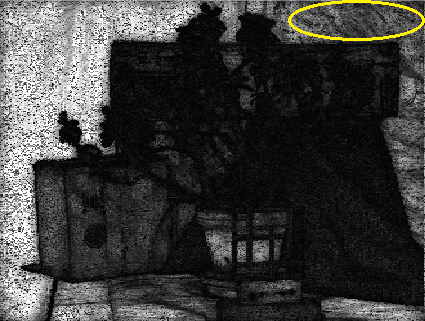}}
	\subfigure[$61 \times 61$]{\includegraphics[width=0.19\linewidth]{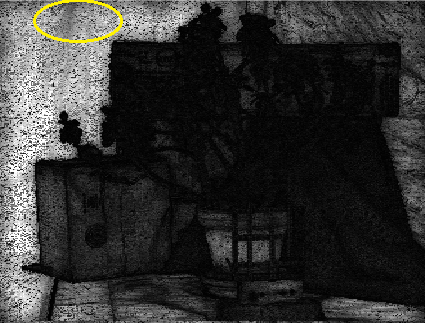}} \\
		\subfigure[Grayscale]{\includegraphics[width=0.19\linewidth]{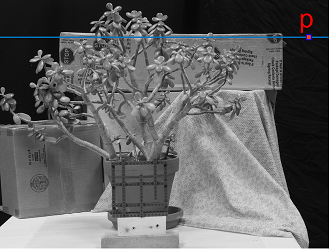}}
	\subfigure[$15 \times 15$]{\includegraphics[width=0.19\linewidth]{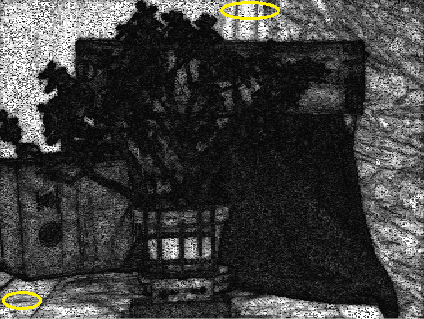}}
		\subfigure[$31 \times 31$]{\includegraphics[width=0.19\linewidth]{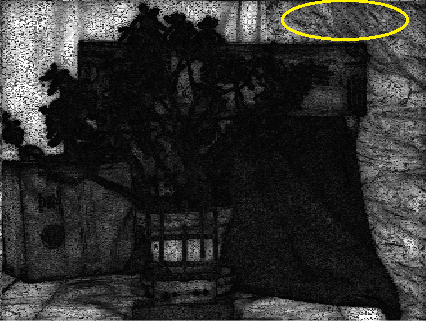}}
	\subfigure[$61 \times 61$]{\includegraphics[width=0.19\linewidth]{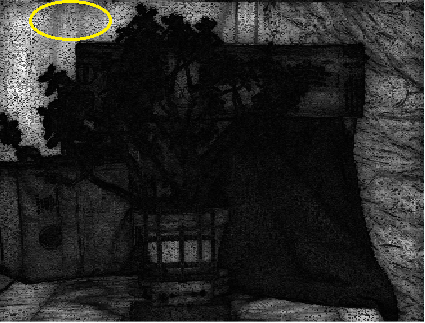}} 
\caption{Results of companion transform under different neighborhood sizes for the ``Jadepl'' dataset (left image on top row and right image on bottom). The transformation results are brightened for better viewing. The results show that companion transform successfully adds distinguishable features to texture-less areas; see regions highlighted.
%\mlc{highlight the regions.}% To involve supportive color information from a wider perspective, a relatively larger window is preferred.
}
\label{fig:companion}
\end{figure*}

%\subsubsection{Companion Transform.}
\subsubsection{Low Texture}

Besides lighting variations, low or devoid of texture poses another challenge for stereo matching. For a given pixel $p$ within texture-less regions, the best way to estimate its disparity is based on its neighbors who have similar depth 
% (on the same object and have similar color as $p$) 
but are in texture-rich areas (have sufficient visual cues for accurate disparity estimation). Traditional stereo algorithms \cite{scharstein2002taxonomy} mostly utilize cost aggregation, segmentation-based matching, or global optimization for disparity computation to handle ambiguous regions. As mentioned above, our intention is to feed the neural networks with global information. Hence, a novel companion transform is designed and applied in the pre-processing step.

The idea of companion transform is inspired by SGM \cite{hirschmuller2008stereo}, which suggests performing smoothness control by minimizing an energy function on 16 directions. In our case, we want to design a transformation that can add distinguishable features to texture-less area.  Hence, for a given pixel $p$, we choose to count the number of pixels that: 1) have the same intensity as $p$ and 2) lie on one of the rays started from $p$. We refer these pixels as $p$'s companions and the transform as companion transform. In practice, we found 8 ray directions (left, right, up, down, and 4 diagonal directions) work well, though other settings (4 or 16 directions) can also be used.
% with exact the Our solution to aggregate global influence is implemented by checking the total number of a pixel's companions on different directions, and those companions have identical color intensity with the central pixel. We here \mwd{practice the transform on eight directions, including two horizontal directions, two vertical directions and four diagonal directions, though it can also be applied on more directions. Selecting all directions are not suggested, as it involves too much local information.} The companion transform $C(p)$ for pixel $p$ can be calculated by: \mlc{do you add the count from all directions together into one number?  if so, why just 8 directions?  Why not count all pixels with the same intensity in the square window?  also, counting only pixels with exact same intensity seems sensitive to noises...}
\begin{equation}
C(p) = \frac{1}{|N_p\cap \Omega_p|}\sum_{q \in N_p \cap \Omega_p}{(I(q) == I(p) ? 1 : 0)} ~,
\label{eq:3_1}
\end{equation}
where $\Omega_p$ is a set containing pixels on the rays started from $p$.

\begin{figure}[h]
\begin{center}
   \includegraphics[width=0.98\linewidth]{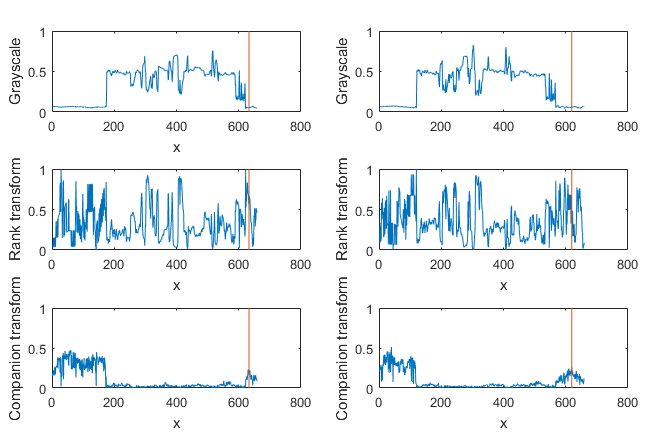} 
\end{center}
\caption{Comparison among information carried in different channels (grayscale, $31 \times 31$ rank transform, and $61 \times 61$ companion transform). The curves are plotted based on the values of different pixels on the same row marked in blue in Figure \ref{fig:companion}. Left side shows the left view, with the position of target pixel $p$ marked by red vertical lines. Right side shows the right view, where the red line shows the position of the correct corresponding pixel of $p$. Due to the lack of textures, neither the grayscale nor the rank transform channels provide distinguishable pattens for matching. The companion transform can amend information that is useful for the matching-CNN.}
\label{fig:curves}
\end{figure}

Figure \ref{fig:companion} shows the results of companion transform under different window sizes. Figure~\ref{fig:curves} further illustrates how the companion transform result adds distinguishable pattens to a pixel in texture-less area. 

\subsubsection{Training Data}

To train our CNN model, the 15 image pairs from Middleburry 2014 stereo training dataset \cite{scharstein2014high}, which contains 
%sufficient 
examples for lighting variations and texture-less areas, are utilized. Each input image is first converted to grayscale before applying rank and companion transforms. The outputs of the two transforms, together with the grayscale images, form multi-channel images. Each training sample contains a pair of image patches centered at pixel $(x, y)$ in left image and $(x-d,y)$ in right image, respectively. The input sample is assembled into a 3D matrix $M[w,w,2\times l]$, where $w$ is the size of the patches and $l$ is the number of channels in the multi-channel image.  The ground-truth disparity values provided by the dataset are used to select matched samples and random disparity values other than the ground truth are used to generate mismatched samples. Similar to \cite{zbontar2016stereo}, we sample matching hypotheses so that the same number of %negative and positive 
matches and mismatches are used for training. 
 The proposed matching-Net is then trained to output value ``0'' for correct matches and  ``1'' for mismatches.

\subsection{Disparity Computation}

For each novel stereo image pair, the matching-Net trained above is used to generate a 3D cost volume $C_s(x,y,d)$, where the value at location $(x,y,d)$ stores the cost of matching pixel $(x,y)$ in the left image with $(x-d,y)$ in the right image.  The higher the value, the more likely the corresponding pair of pixels are mismatches since the network is trained to output ``1'' for mismatches. Unlike many existing approaches that resort to complex and heuristically designed cost aggregation and disparity optimization approaches \cite{scharstein2002taxonomy}, here we rely on the learning network to distinguish correct matches from mismatches. Expecting the correct matches to have the smallest values in the cost volume $C_s(x,y,d)$, the simplest WTA optimization is applied to compute the raw disparity map.

\begin{equation}
D_e(x,y)=\arg\min_d C_s(x,y,d) ~.
\label{eq:3_1}
\end{equation}

\subsection{Evaluation-Net}

The matching-net is trained to measure how well two images patches, one from each stereo image, match. It makes decision locally and does not check the consistency among best matches found for neighboring pixels. When the raw disparity maps are computed by local WTA, they inevitably contain mismatches, especially in occluded and low-textured areas. 
%Simply selecting disparities depending on matching scores computed by WTA may exclude low-textured areas but maintain noise, since the latter may achieve more convincing matching scores.} 
To filter out these mismatches, we construct another CNN model, evaluation-Net, to implement consistency check and perform confidence measure. 

\begin{figure}[t]
\centering
\includegraphics[width=0.6\linewidth]{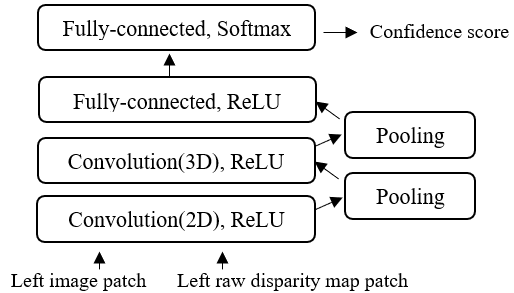}
\caption{Architecture used for the evaluation-Net. The image patches here are generally bigger than the ones used in the matching-Net. Therefore, multiple pooling layers are added for efficiency. Detailed model configuration can be found in Section \ref{sec}.}
\label{fig:snet}
\end{figure}

Learning-based confidence measures have been successfully applied on detecting mismatches and further improving the accuracy of stereo matching \cite{poggi2017quantitative}. Similar to the 2D CNN model for error detection proposed in \cite{ye2017efficient}, only left images and their disparity maps are selected to train our model.  A key difference, however, is that no handcrafted operation is involved in our approach to fuse left and right disparity maps. In addition, the network contains both 2D and 3D convolutional layers to effectively identify mismatches from disparity maps; see Figure \ref{fig:snet}. 3D convolution is adopted here to allow the network learn from the correlation between pixels' intensities and disparity values.

\begin{figure}[t]
\centering
	\subfigure[raw disparity]{\includegraphics[width=0.32\linewidth]{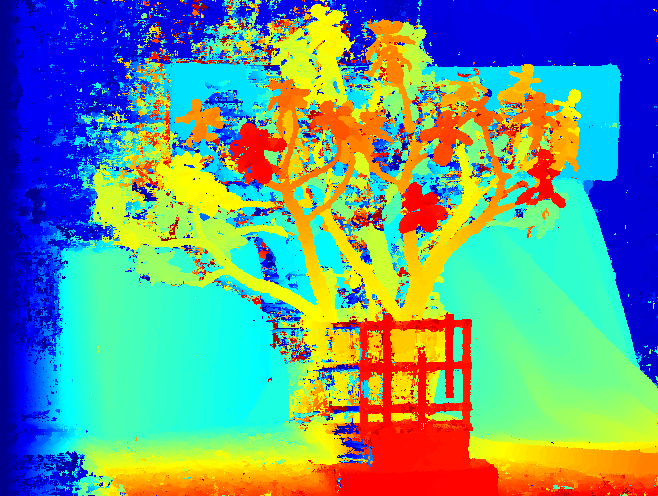}}
	\subfigure[mismatches]{\includegraphics[width=0.32\linewidth]{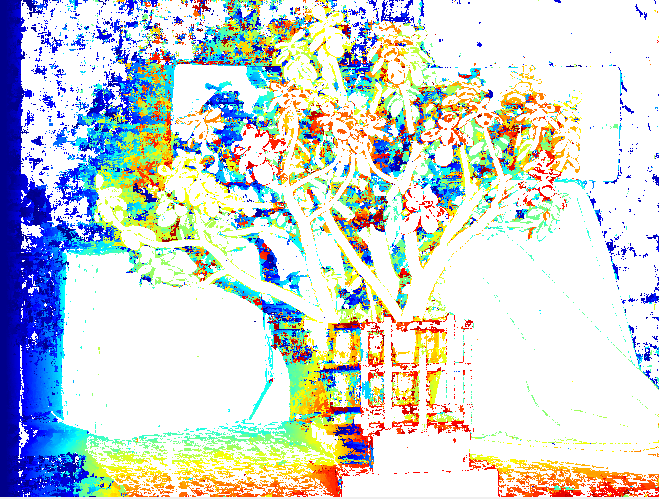}}
	\subfigure[matches]{\includegraphics[width=0.32\linewidth]{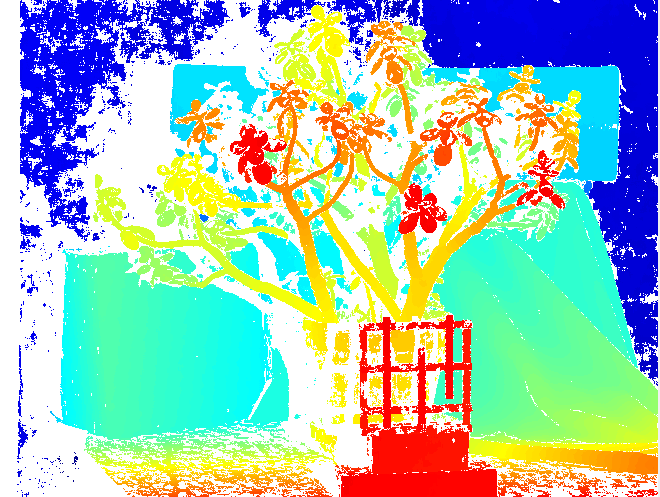}}
\caption{Training samples. Pixels in a given disparity map (a) is classified into mismatches (b) and accurate matches (c) using ground-truth disparity.}
\label{fig:select_samples}
\end{figure}

The evaluation-Net is trained using both matches and mismatches in the estimated disparity maps $D_e(x,y)$ for all training images. Mismatches are identified by comparing $D_e(x,y)$ with ground-truth disparity maps $D_t(x,y)$. Here, a pixel $(x,y)$ is considered as mismatched \textit{iff.}
\begin{equation}
\|D_e(x,y)-D_t(x,y)\| > T_e ~, 
\label{eq:3_2}
\end{equation}
where  $T_e$ is a threshold value commonly assigned with $1$ pixel; see Figure~\ref{fig:select_samples}(b-c). 

In the estimated disparity map $D_e(x,y)$, the majority pixels have correct disparity values, resulting in much more positive (accurately matched) samples
than negative (mismatch) samples.  Hence, we collect and use all negative samples and randomly generate the same number of positive samples. 
For each selected sample $(x,y)$, we extract grayscale and estimated disparity values from patches centered at $(x,y)$ to form a $w' \times w' \times 2$ matrix. The evaluation-Net is then trained to output value ``0" for negative samples and ``1" for positive samples. The output of the evaluation-Net can then be used to filter out potential mismatches which achieve scores lower than a confidence threshold $R$.

\section{Experimental Results} \label{sec}

In this section, we present the ``hyperparameters'' \cite{zbontar2016stereo} for both of the proposed CNN models, which are followed by a set of performance evaluations.  The goal of the evaluations is to find out: 1) whether the non-parametric transforms can help improving the disparity map accuracy generated using the matching-Net; and 2) how well the overall dual-CNN approach performs compared to the state-of-the-art sparse stereo matching techniques.
%does evaluation-Net detect mismatches as a confidence measure. 

\begin{table*}[t]
\caption{Hyperparameters of the matching-Net and evaluation-Net. Here, ``Conv", ``Mp" and ``Fc" denote convolutional layer, max pooling layer, and fully connected layers respectively. %Within Conv1 and Conv2 of matching-Net, we additionally set their padding size as $1 \times 1 \times 1$.
}
\begin{center}
%\begin{tabular}{ |c|c|c|c|c|c| }
\begin{tabular}{ |c c c c c c| }
  \hline
	%\cline{1-3} \cline{4-6}
	\multicolumn{3}{|c}{ matching-Net} & \multicolumn{3}{c|}{ evaluation-Net}\\
	%\hline
	Attributes & Kernel size, quantity  &  ~~Stride  size & Attributes & Kernel size, quantity &  ~~Stride size\\
  \hline
	\hline
	Input & \multicolumn{2}{c}{ $11 \times 11 \times 6$, 1 } & Input & \multicolumn{2}{c|}{ $101 \times 101 \times 2$, 1 }\\
	%\hline	
	Conv1(2D) & $~~3 \times 3 \times 1$, 32 &  $~~1 \times 1 \times 1$  &~~Conv1(2D) &  $~~3 \times 3 \times 1$, 16 &  $~~1 \times 1 \times 1$ \\
	%\hline	
	Conv2(3D) & $~~~3 \times 3 \times 2$, 128 &  $~~1 \times 1 \times 2$ &~~Mp1 &  $~~2 \times 2 \times 1$, 16 &  $~~2 \times 2 \times 1$ \\
  %\hline
	Conv3(3D) & $~~3 \times 3 \times 3$, 64 &  $~~1 \times 1 \times 1$ &~~Conv2(2D) &  $~~3 \times 3 \times 1$, 32 &  $~~1 \times 1 \times 1$ \\
  %\hline  
	FC1 & \multicolumn{2}{c}{ 1600 } &~~Mp2  &  $~~2 \times 2 \times 1$, 32 &  $~~2 \times 2 \times 1$ \\
  %\hline  
	FC2 & \multicolumn{2}{c}{ 128 } &~~Conv3(2D) &  $~~3 \times 3 \times 1$, 64 &  $~~1 \times 1 \times 1$ \\
  %\hline
	Output & \multicolumn{2}{c}{2} &~~Mp3  &  $~~2 \times 2 \times 1$, 64 &  $~~2 \times 2 \times 1$ \\
	  %\hline
	- & \multicolumn{2}{c}{-} &~~Conv4(3D) &  $~~3 \times 3 \times 2$, 128 &  $~~1 \times 1 \times 1$ \\
	  %\hline
	- & \multicolumn{2}{c}{-} &~~Mp4  &  $~~2 \times 2 \times 1$, 128 &  $~~2 \times 2 \times 1$ \\
	%\hline
	- & \multicolumn{2}{c}{-} &~~FC1 &  \multicolumn{2}{c|}{ 128} \\
	%\hline
	- & \multicolumn{2}{c}{-} &~~Output & \multicolumn{2}{c|}{ 2 }\\
	
	\hline	
\end{tabular}
\end{center}
\label{table:net}
\end{table*}

\textbf{Hyperparameters and implementations:}
The input of the matching-Net is a 3D matrix that consists of $l=6$ layers in our experiment. Both left and right images contains $3$ layers, including the grayscale image, a rank transform ($w_r=31$), and a companion transform ($w_c=61$) respectively. 
Different layers from the left and right images are stored in the matrix in alternating order. For the evaluation-Net, the input contains only two layers of data: one is the grayscale image and the other the raw disparity map, both from the left image. 
Table \ref{table:net} shows the hyperparameters of our experimental models.

The implementation of our CNN models are based on Tensorflow using classification cross-entropy loss, $-(t\log{s}+(1-t)\log(1-s))$, where $s$ denotes the output value. Here, we set $t=1$ for mismatches and $t=0$ for matches to train the matching-Net as in ``MC-CNN-acrt", but $t=1$ for positive samples and $t=0$ for negative samples to perform confidence measure through the evaluation-Net. Both models utilize a gradually decreasing learning rate from 0.002 to 0.0001, and arrive a stable state after running 20 epochs on full training data. %In terms of runtime optimization, interested readers are referred to \cite{zbontar2016stereo}.

\begin{figure}[t]
\centering
	\subfigure[]{\includegraphics[height=0.52\linewidth]{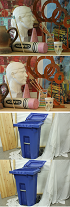}}
	\subfigure[]{\includegraphics[height=0.52\linewidth]{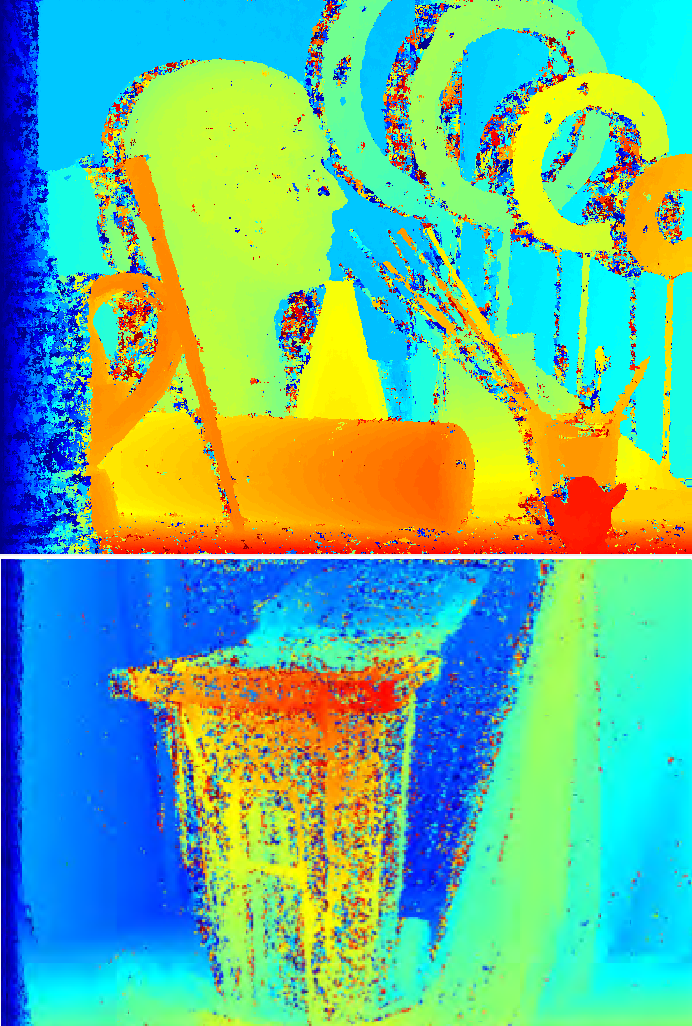}}
	\subfigure[]{\includegraphics[height=0.52\linewidth]{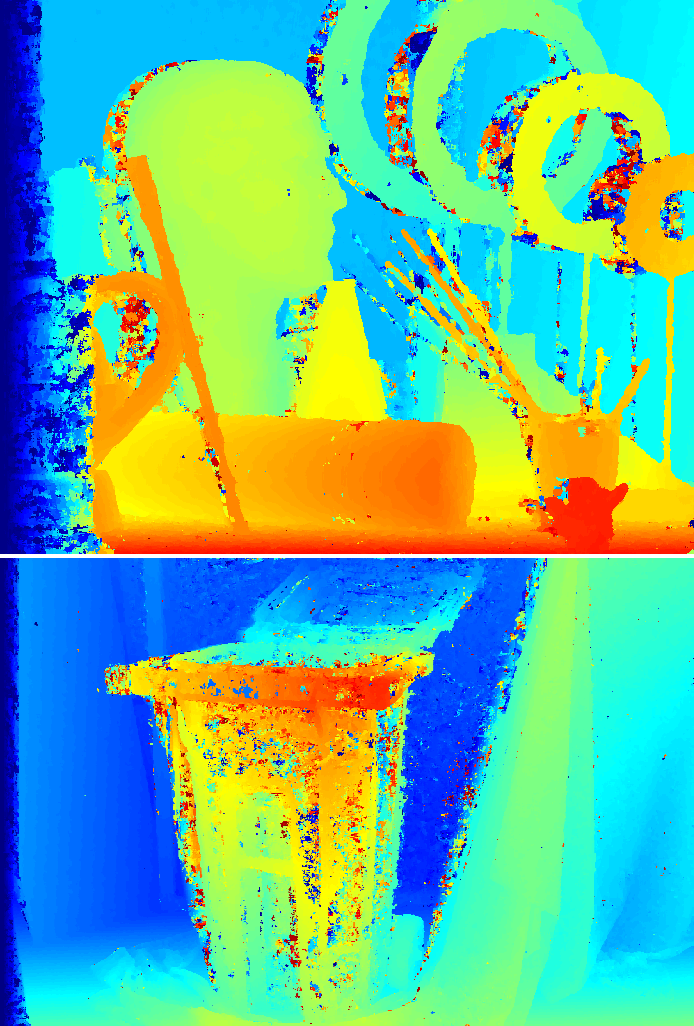}}
\caption{Comparison on dense disparity maps generated by ``MC-CNN-acrt'' (b) and matching-Net (c). Top stereo image pair (``ArtL'') contains lighting condition changes, whereas the bottom one (``Recyc'') contains areas with low texture. Thanks to the rank and companion transforms, the disparity maps generated by our approach are much smoother and have fewer mismatches.}

\label{fig:raw_results}
\end{figure}

\textbf{Effectiveness of non-parametric transforms:}
The overall structures of ``MC-CNN-acrt" and matching-Net are quite similar.  The key difference is that the input patches of ``MC-CNN-acrt" are grayscale images only, whereas our matching-Net uses additional non-parametric transforms. Hence, to evaluate the effectiveness of non-parametric transforms, we here compare the raw disparity maps generated by the two approaches.
Based on the same training dataset from Middlebury \cite{scharstein2014high}, 
Figure \ref{fig:raw_results} visually compares the raw disparity maps generated by ``MC-CNN-acrt" and matching-Net. It suggests that the additional transforms allow the network to better handle challenging cases. Our raw disparity maps achieves $18.69\%$ compared to $22.91\%$ of ``MC-CNN-acrt" regarding the mean percentage error (MPE) (over $1$-pixel difference for half resolution) of non-occlusion areas.

%\subsection{RMS Comparison}

\begin{figure*}[h]
\begin{center}
\subfigure[training]{ \includegraphics[height=0.32\linewidth]{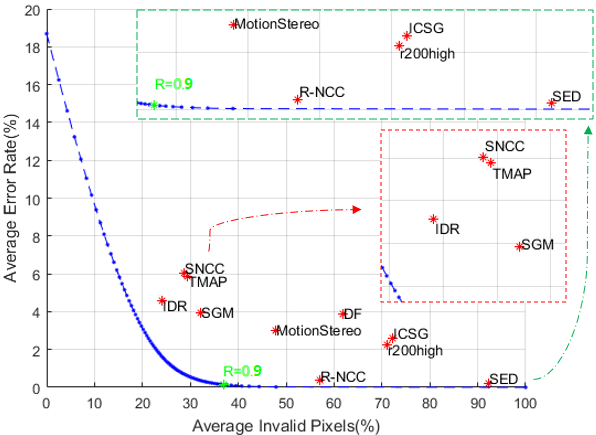} }
\subfigure[testing]{ \includegraphics[height=0.32\linewidth]{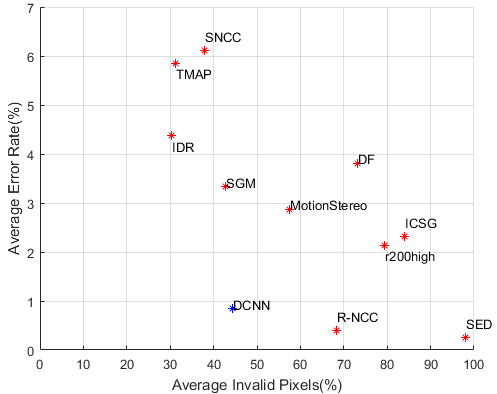} }
\end{center}
\caption{
Comparison with the top ten approaches on the Middlebury Stereo Evaluation site \cite{scharstein2014high}: SED \cite{sed}, R-NCC (unpublished work), r200high \cite{keselman2017}, ICSG \cite{shahbazi2016revisiting}, SGM \cite{hirschmuller2008stereo}, DF (unpublished work), MotionStereo (unpublished work), IDR \cite{6213098}, TMAP \cite{psota2015map} and SNCC \cite{einecke2010two}. Performances of different approaches on both training (a) and testing (b) datasets are plotted on non-occlusion error rates v.s. invalid pixels rates plot.  The relative position of these approaches on the two datasets are similar.  On training datasets, where the ground truth disparity maps are available, we show the performance of our approach under different confidence threshold $R$ settings as a curve.%  \mlc{use blue color for DCNN in (b).  also the x-axis should starts from 0!}
}
\label{fig:err_inv}
\end{figure*}

\begin{figure*}[t]
\begin{center}

  \includegraphics[width=0.124\linewidth]{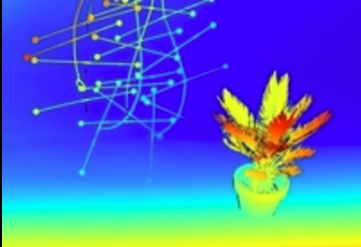}\includegraphics[width=0.124\linewidth]{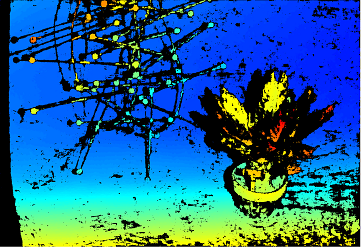}\includegraphics[width=0.124\linewidth]{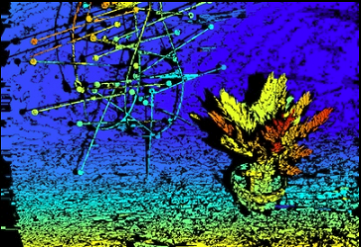}\includegraphics[width=0.124\linewidth]{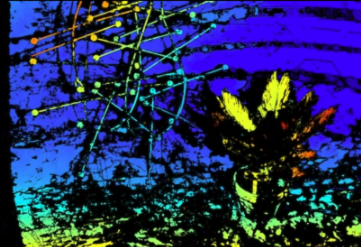}\includegraphics[width=0.124\linewidth]{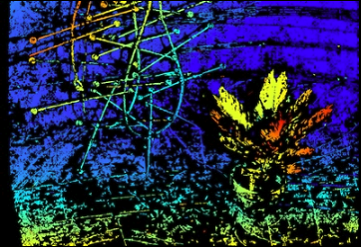}\includegraphics[width=0.124\linewidth]{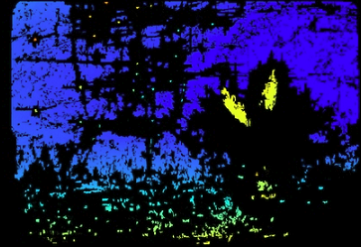}\includegraphics[width=0.124\linewidth]{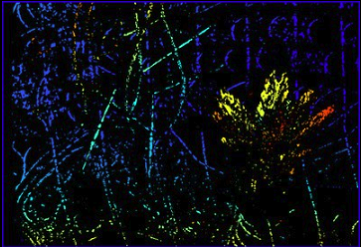}\includegraphics[width=0.124\linewidth]{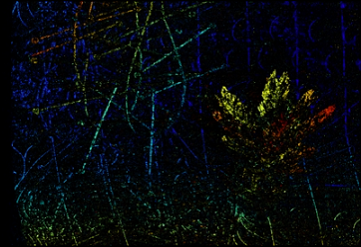}
	\vspace{.01in}
	\includegraphics[width=0.124\linewidth]{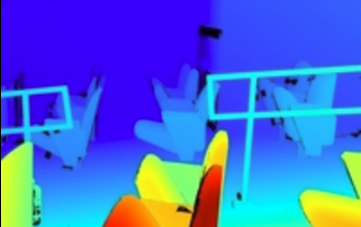}\includegraphics[width=0.124\linewidth]{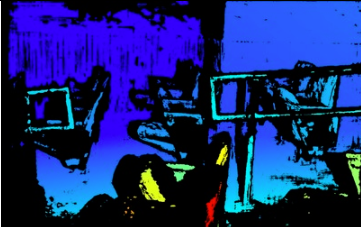}\includegraphics[width=0.124\linewidth]{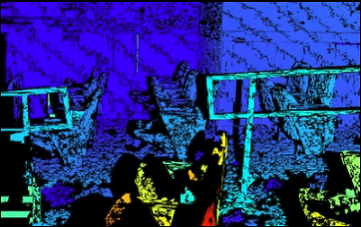}\includegraphics[width=0.124\linewidth]{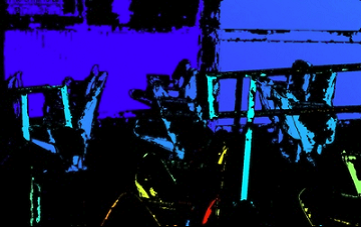}\includegraphics[width=0.124\linewidth]{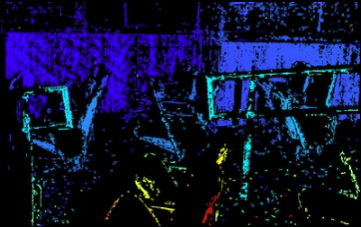}\includegraphics[width=0.124\linewidth]{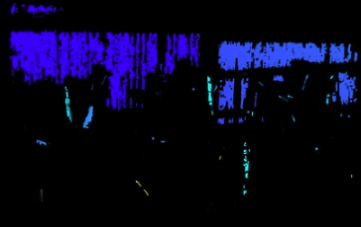}\includegraphics[width=0.124\linewidth]{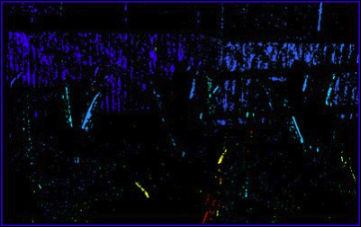}\includegraphics[width=0.124\linewidth]{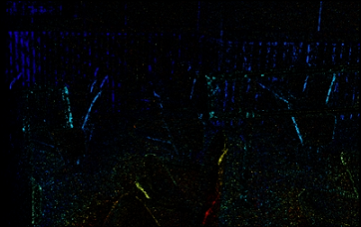}
	 %\includegraphics[width=0.124\linewidth]{figures/test2_SNCC}\includegraphics[width=0.124\linewidth]{figures/test6_SNCC}\includegraphics[width=0.124\linewidth]{figures/test8_SNCC}
	
	%\vspace{.05in}
	
\end{center}
   \caption{Comparison of sparse disparity maps regarding ``Austr" and ``ClassE" (with lighting variation) of the testing dataset on the Middlebury Stereo Evaluation site \cite{scharstein2014high}.  First column shows the ground truth, and columns 2 to 8 are the disparity maps generated by DCNN, TMAP~\cite{psota2015map}, IDR \cite{6213098}, 
%SNCC~\cite{einecke2010two}, 
SGM~\cite{hirschmuller2008stereo}, R-NCC (unpublished work), r200high~\cite{keselman2017} and ICSG~\cite{shahbazi2016revisiting} respectively.} %\mlc{missing an algorithm here?}

\label{fig:results}
\end{figure*}

\textbf{Comparison with sparse stereo matching approaches:}
Almost all state-of-the-art sparse stereo matching approaches have submitted their results to Middlebury evaluation site \cite{scharstein2014high}. 
%Hence, we here train our dual-CNN approach on their training datasets as well. 
Our approach (referred as ``DCNN") on ``test sparse" currently ranks the $3^{rd}$ under the ``bad 2.0'' category. We would like to emphasize that simply comparing error rates of sparse disparities maps does not offer the whole picture on algorithm performance as it favors approaches that output fewer disparity values (a.k.a. more invalid pixels). For a fair comparison, a non-occlusion error rates v.s. invalid pixels rates plot is used to show the performance of different approaches on both the training and testing datasets; see Figure \ref{fig:err_inv}.  The comparison suggests that our approach under $R=0.9$ setting provides a very good balance between output disparity density and disparity accuracy.  In addition, the plot on the training dataset also shows that, under the same output disparity density, our approach provides lower non-occlusion error rates than existing approaches.
Figure \ref{fig:results} further visually compares the disparity maps generated by different approaches.

The root-mean-square (RMS) metric \cite{scharstein2014high} is also used here for evaluation.
%\begin{equation}
%rms = \sqrt{\frac{1}{|D_s|} \sum_{(x, y ) \in D_s}{(D_e(x,y)-D_t(x,y))^2}} ~,
%\label{eq:4_1}
%\end{equation}
%where $D_s$ denotes all pixels with disparity values reported in $D_e$ and having ground-truth disparity values in $D_t$.
%%, and $n_s$ is the total number \mlc{what is total number?}
Since square errors are used, the RMS metric provides stronger penalization to large disparity errors than the average absolute error (``avgerr'') metric. 
Our approach on the testing dataset currently ranks on the top under the ``rms" category; see Table \ref{table:RMS}.
\begin{table}[h]
\caption{Comparisons of the state-of-the-art approaches under the RMS metric.}
\begin{center}
\begin{tabular}{ |c c c c| }
\hline
Name & RMS & Name & RMS \\
\hline\hline
DCNN & $3.86_{1}$&MPSV~\cite{bricola:hal-01330139}&$9.25_{4}$\\
R-NCC(unpublished) & $4.61_{2}$&INTS~\cite{huang2016image}& $10.6_{5}$\\
IDR~\cite{6213098}&$8.07_{3}$& SGM~\cite{hirschmuller2008stereo}&$10.9_{6}$\\ 
%MotionStereo & $11.3_{7}$ TMAP~\cite{psota2015map}& $11.5_{8}$\\
\hline	
\end{tabular}
\end{center}
\label{table:RMS}
\end{table}

%for training and test datasets, respectively; see Figure \ref{fig:train_test_rms}. 

%\begin{figure}[h]
%\begin{center}
%\subfigure[]{\includegraphics[width=0.7\linewidth]{figures/train_rms}}
%\subfigure[]{\includegraphics[width=0.7\linewidth]{figures/test_rms} }

%\end{center}
%\caption{Comparison of training dataset (a) and test dataset (b) with other semi-dense stereo results regarding RMS.}
%\label{fig:train_test_rms}
%\end{figure}

%\begin{figure}[t]
%\centering
%	\subfigure[``MotorE	"]{\includegraphics[height=0.19\linewidth]{figures/AUC_train6}}
%	\subfigure[``Jadepl"]{\includegraphics[height=0.19\linewidth]{figures/AUC_train4}}
%	\subfigure[``Recyc"]{\includegraphics[height=0.19\linewidth]{figures/AUC_train13}}
%\caption{\ml{Performances on different image pairs. Our approach can effectively select correct matches from raw disparity maps and the corresponding curves are very close to the %optimal ones.}} %\mlc{AUC is an area (single number), not a curve...}}
%\label{fig:auc}
%\end{figure}

\textbf{AUC evaluation:}
The Area Under the Curve (AUC) metric introduced by Hu and Mordohai \cite{hu2012quantitative} has been used as a metric for evaluating various confidence measures over the past few years. It measures how effectively the confidence measures can filter out mismatches under different parameter settings, rather than only checking the performance under one set of parameters. Since a large set of sparse disparity maps need to be evaluated, this measure can only be computed on datasets with published ground truth.  Following the practice in \cite{tosi2017learning}, we train our dual-CNN approach only on the 13 additional image pairs with ground truth from Middlebury \cite{scharstein2014high} and then test it on the 15 training image pairs. Our approach achieves a competitive mean AUC value $0.0522$ compared to 0.0728, 0.0680 and 0.0637 attained respectively by the state-of-the-art approaches APKR \cite{kim2014stereo}, O1 \cite{poggi2016learning} and CCNN \cite{Poggi2016LearningFS} reported in \cite{tosi2017learning}, which compares various confidence measures on the raw disparity maps from \cite{zbontar2016stereo}.

\section{Conclusion}

A novel learning based semi-dense stereo matching algorithm is presented in this paper. The algorithm employs two CNN models.  The first CNN model evaluates how well two image patches match.  It serves the same purpose as ``MC-CNN-acrt'', but takes additional rank and companion transforms as input.  These two transforms introduce global information and distinguishable patterns into the network; and hence areas with lighting changes and/or lack of textures can be more accurately matched.  As a result, the optimal disparity values can be computed using the simplest WTA optimization. No complicated global disparity optimization algorithms or additional post-processing steps are required.
The second CNN model is used for evaluating the disparity values generated and filter out mismatches. Taking only one of the stereo images and the disparity map as input, the evaluation-Net can effectively label mismatches, without the needs for heuristically designed process such as left-right consistency check and median filtering.

Our work suggests that, once sufficient information is fed to the network, CNN-based models can effectively predict the correct matches and detect mismatches. For the future work, we plan to investigate how to reduce the training and labeling costs so that the algorithm can be applied to real-time applications. We also plan to apply the algorithm to multi-view stereo matching for  3D reconstruction applications.

{\small
\bibliographystyle{ieee}
\bibliography{egbib}
}

\end{document}